\documentclass[lettersize,journal]{IEEEtran}
\usepackage{amsmath,amsfonts}
\usepackage{algorithmic}
\usepackage{algorithm}
\usepackage{array}
\usepackage[caption=false,font=normalsize,labelfont=sf,textfont=sf]{subfig}
\usepackage{textcomp}
\usepackage{stfloats}
\usepackage{url}
\usepackage{verbatim}
\usepackage{graphicx}
\usepackage{cite}
\usepackage{multirow}
\usepackage{color}
\usepackage[colorlinks, linkcolor=black, anchorcolor=blue, citecolor=black]{hyperref}
\begin{document}

\title{Beyond Graph Model: Reliable VLM Fine-Tuning via Random Graph Adapter}

\author{Bo~Jiang, Xueyang~Ze, Beibei~Wang, Xixi~Wang, Xixi~Wan,
        and~Bin~Luo,~\IEEEmembership{Senior Member,~IEEE}
\thanks{The authors are with the Anhui Provincial Key Laboratory of Multimodal Cognitive Computation, School of Computer Science and Technology, Anhui University, Hefei 230601, China.}
}

\markboth{Journal of \LaTeX\ Class Files,~Vol.~14, No.~8, August~2021}%
{Shell \MakeLowercase{\textit{et al.}}: A Sample Article Using IEEEtran.cls for IEEE Journals}

\IEEEpubid{0000--0000/00\$00.00~\copyright~2021 IEEE}

\maketitle

\begin{abstract}
Textual adapter-based tuning methods have shown significant potential in transferring knowledge from pre-trained Vision-Language Models (VLMs) to downstream tasks. 
Existing works generally employ the deterministic textual feature adapter to refine each category textual representation. 
However, due to inherent factors such as different attributes and contexts, there exists significant diversity in textual descriptions for each category. Such description diversity offers rich discriminative semantic knowledge 
that can benefit downstream visual learning tasks.
Obviously, traditional deterministic adapter model cannot adequately capture this varied semantic information. 
Also, it is desirable to exploit the inter-class relationships in VLM adapter. 
To address these issues, we propose to exploit \emph{random graph model} into VLM adapter and
develop a novel Vertex Random Graph Adapter (VRGAdapter). 
VRGAdapter 
first models the inherent diverse descriptions of each category and inter-class relationships of different categories simultaneously by leveraging a Vertex Random Knowledge Graph (VRKG) model. 
Then, it employs probabilistic message propagation on VRKG to learn context-aware distribution representation for each class node. 
Finally, it adopts a reparameterized sampling function to achieve textual adapter learning.  
Note that, VRGAdapter provides a more general adapter solution that encompasses traditional graph-based adapter as a special case. 
In addition, to enable more robust performance for downstream tasks, we also introduce a new Uncertainty-guided Multi-branch Fusion (UMF) scheme that dynamically integrates multiple pre-trained models for ensemble prediction.  
Extensive experiments on multiple benchmark datasets demonstrate the effectiveness of our approach.

\end{abstract}

\begin{figure}[htbp]
    \centering
    \includegraphics[width=\linewidth]{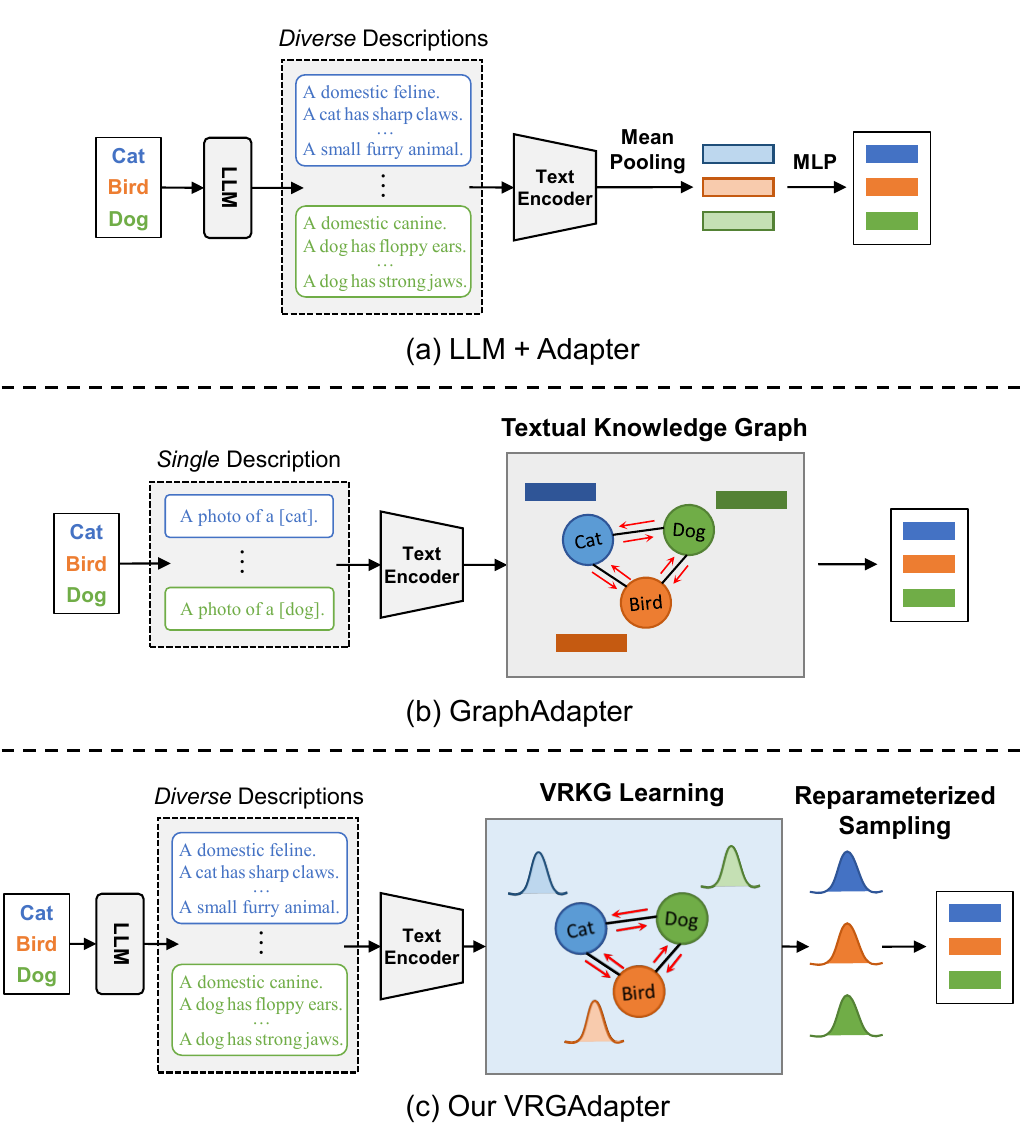}
    \caption{Comparison of different textual branch adaptation strategies for VLMs.
(a) LLM + Adapter: Leverages LLM to generate diverse class descriptions~\cite{cupl}, which are encoded via a text encoder and aggregated by mean pooling. The mean features are then refined by a lightweight MLP adapter. 
(b) GraphAdapter~\cite{graphadapter}: Uses handcrafted prompts to construct a knowledge graph with textual and visual nodes. Here, we only show the textual graph, where node features are fixed and enhanced via GCN to refine class representations. 
(c) VRGAdapter: Generates multiple semantic descriptions per class using LLM and constructs a vertex random knowledge graph to capture both intra-class semantic diversity and inter-class relationships. 
The message propagation is then conducted on VRKG to learn context-aware distribution representations. The sampling technique is finally adopted for final textual adaptation.}

    \label{fig:intro}
\end{figure}

\section{Introduction}
\label{sec:intro}
\IEEEPARstart{V}{ision}-Language Models (VLMs), such as CLIP~\cite{clip} and BLIP~\cite{Blip,Blip-2}, have demonstrated remarkable success in various downstream tasks by leveraging knowledge learned from large-scale image-text pairs.
Among these, VLMs effectively transfer semantic knowledge via textual prompts like ``a photo of a [class]", enabling superior performance in different tasks, including zero-shot generalization~\cite{calip,sus}, few-shot classification~\cite{coop}, out-of-distribution  detection~\cite{detection}, and semantic segmentation~\cite{segmentation}. 
In recent years, numerous efforts have been made to enhance the transfer learning capabilities of VLMs under a limited amount of trainable data.
Broadly speaking, existing transfer learning strategies for tuning VLMs on downstream tasks can be categorized into three types: full fine-tuning~\cite{fullft1,FD-align}, prompt tuning~\cite{coop,cocoop,ProDA,yao2025bi}, and adapter-based tuning~\cite{clip-adapter,tip,taskres,graphadapter,ClusterAdapter,Mma}.
Full fine-tuning methods focus on updating all parameters of the pre-trained model but require substantial computational resources and risk overfitting in low-data scenarios~\cite{fullft1,FD-align}. 
Prompt-tuning methods introduce learnable textual prompts to modulate VLMs' behavior and adapt their semantic understanding for downstream tasks~\cite{coop,maple,BayesianPrompt,promptKD}. However, these methods require computing gradients of the text encoder during training to optimize the learnable prompts, usually leading to high computational costs.
\IEEEpubidadjcol  
Adapter-based tuning methods focus on attaching lightweight modules 
to the outputs of frozen pre-trained textual or visual encoders, enabling feature refinement for downstream tasks while preserving the original VLM knowledge~\cite{clip-adapter,tip,taskres,ClusterAdapter,cafo}.
For example, CLIP-Adapter~\cite{clip-adapter} proposes a trainable MLP adapter following the pre-trained CLIP encoder to enhance feature representation.
TaskRes~\cite{taskres} introduces residual vectors with textual features to learn task-specific knowledge.  
GraphAdapter~\cite{graphadapter} constructs a dual knowledge graph and uses the Graph Convolution Network (GCN) to model inter-class relationships, but treats each class node as a deterministic vector representation.
TextRefiner~\cite{TextRefiner} leverages the internal knowledge to provide fine-grained information and refine text prompts.

After reviewing the above existing works, we observe that 
{existing textual adapters generally employ the deterministic textual feature function to refine the textual representation for each class/category. }
However, due to inherent factors such as different attributes and contexts, there exists significant diversity of textual descriptions for each category~\cite{cupl}. 
For instance, the class `cat' can be described in text with various phrases/sentences such as \{`A domestic feline', `A small furry animal with whiskers', `A cat has sharp claws and erect ears'\}. 
Obviously, \emph{such description diversity/variation provides 
rich discriminative information
for visual learning which should be fully explored for fine-tuning VLMs on the downstream visual learning tasks}, such as zero-shot~\cite{cupl} and few-shot image classification~\cite{cafo}.   
However, existing works generally either adopt a single description~\cite{taskres,Mma} 
or employ the `mean' representation of some  descriptions~\cite{cafo,LwEIB,cupl} for feeding the textual adapter which \textbf{obviously {fail to adequately capture the description variation of each category in the adapter,  thereby leading to sub-optimal solution on downstream tasks.}}
Also, it is desirable to further exploit the inter-class relationships to enhance the class representation in VLM adapter~\cite{graphadapter,HeGraphAdapter}.

Based on the above observations, in this paper, 
  we propose to exploit \textbf{random graph learning model into VLM adapter} and develop a novel Vertex Random Graph Adapter (VRGAdapter) for fine-tuning VLMs. 
As illustrated in Fig.~\ref{fig:intro}, 
our  VRGAdapter leverages a \textbf{V}ertex \textbf{R}andom \textbf{K}nowledge \textbf{G}raph (VRKG) learning model to fully capture the knowledge of both semantic variation and inter-class relationships to adapt the category textual representation for the downstream tasks. 
Specifically, VRGAdapter involves three components. 
\emph{First}, the 
{V}ertex {R}andom {K}nowledge {G}raph (VRKG)  is  constructed with each node denoting a class represented as a Gaussian distribution. Each distribution is initialized by using $M$ diverse textual descriptions generated by Large Language Models (LLMs)~\cite{gpt3}, thereby inherently encapsulating the semantic variability within each class. 
Each edge of VRKG encodes the inter-class dependency of different classes. 
\emph{Then}, we employ a message propagation mechanism on VRKG to learn a context-aware distribution representation for each class. 
\emph{Third}, a 
reparameterized sampling function is adopted to achieve textual feature representation for adapter modulation. 
Note that, the proposed VRGAdapter provides a general differentiable solution for fine-tuning VLMs on the downstream tasks. 
Comparing with traditional graph-based adapter (termed GraphAdapter~\cite{graphadapter}), our VRGAdapter leverages significantly more discriminating semantic knowledge in the adapter through a probabilistic model. 
Moreover, GraphAdapter~\cite{graphadapter} can be regarded as a specific case of our VRGAdapter with diversity $M=1$. 
We will show that our VRGAdapter can significantly outperform GraphAdapter~\cite{graphadapter} approach in many tasks. 

In addition, some recent works~\cite{cafo,amu,BayesianEnsemble} propose to integrate multiple pre-trained models like MoCo~\cite{mocov3} and DINO~\cite{dino} to obtain better performance for the downstream tasks.  
However, how to effectively integrate multiple pre-trained models is still an open problem. 
To address this problem, 
we introduce an Uncertainty-guided Multi-branch Fusion (UMF) scheme that dynamically integrates predictions from multiple pre-trained foundation models. 
Unlike static ensemble approaches, our approach quantifies model-specific uncertainty for each input sample by leveraging Kurtosis~\cite{kurtosis}, a statistical metric that measures prediction distribution sharpness. 
Model demonstrating higher confidence receives larger weight in the fusion process, enabling each sample to adaptively integrate predictions from multiple pre-trained models.

Overall, the main contributions of this paper can be summarized as follows:
\begin{itemize}
    \item We propose to exploit the random graph representation model into the VLMs adapter and present a novel Vertex Random Graph Adapter (VRGAdapter) for fine-tuning VLMs on downstream tasks. 
VRGAdapter provides a general solution for fine-tuning VLMs which {fully captures the rich diversity of each category representation and the inter-class relationships of different categories simultaneously for textual adapter learning}. 
    \item We design a novel Uncertainty-guided Multi-branch Fusion (UMF) scheme to dynamically integrate multiple pre-trained visual models to obtain more reliable learning for the downstream tasks. 
    \item Extensive experiments on 11 benchmark datasets show that the proposed method consistently outperforms existing approaches, validating its superior effectiveness and performance on various downstream tasks.
\end{itemize}

\section{Related Work}
\subsection{Tuning for Vision-Language Models}
Vision-Language Models (VLMs) have demonstrated remarkable capabilities through pre-training on large-scale image-text pairs~\cite{clip,Blip}. Existing works explore different approaches to efficiently transfer VLMs' knowledge to downstream tasks, which can be broadly categorized into three groups: full fine-tuning~\cite{fullft1,FD-align}, prompt tuning~\cite{coop,cocoop,ProDA}, and adapter-based tuning~\cite{clip-adapter,tip,taskres,graphadapter,ClusterAdapter,Mma}.
Full fine-tuning methods~\cite{fullft1,FD-align} update all parameters of the pre-trained model but require substantial computational resources and risk overfitting in low-data scenarios. 
WiSE-FT~\cite{fullft1} first fine-tunes the model weights on a specific dataset, then ensembles them with the zero-shot model weights to improve performance under distribution shifts.
FD-Align~\cite{FD-align} fine-tunes the entire CLIP visual encoder using category-agnostic text as spurious prototypes, while preserving feature distribution consistency during tuning.
Prompt tuning methods~\cite{coop,cocoop,maple} focus on introducing learnable prompts to modulate VLMs' behavior for downstream tasks. For instance,
CoOp~\cite{coop} introduces learnable context vectors to generate task-specific prompts, while CoCoOp~\cite{cocoop} further proposes an instance-conditional prompt learning strategy. 
MaPLe~\cite{maple} adopts a hierarchical prompt learning strategy to enhance CLIP's adaptability by jointly adjusting representations in text and image encoders.
PromptKD~\cite{promptKD} introduces a prompt-based distillation approach that enables a lightweight CLIP model to learn from its larger version using unlabeled domain data.
Adapter-based tuning methods~\cite{clip-adapter,tip,cafo} construct lightweight modules to refine the output features while keeping the pre-trained model frozen. For example,
CLIP-Adapter~\cite{clip-adapter} introduces MLP-based modules to adjust both textual and visual features. 
Tip-Adapter~\cite{tip} constructs a key-value cache to store and retrieve task-specific knowledge.
TaskRes~\cite{taskres} introduces task residual parameters on the text side to learn task-specific knowledge while keeping the original classifier weights frozen. 
GraphAdapter~\cite{graphadapter} leverages the text features of each class to build a knowledge graph and uses Graph Convolutional Network (GCN) to optimize node features.
CaFo~\cite{cafo} proposes a cascade of foundation models, using GPT-3~\cite{gpt3} and DALL-E~\cite{dalle} for textual and visual data augmentation, and integrating prior knowledge from CLIP~\cite{clip} and DINO~\cite{dino} based on similarity.
AMU-Tuning~\cite{amu}  incorporates auxiliary features generated from other large pre-trained models to learn an effective logit bias. 
ClusterAdapter~\cite{ClusterAdapter} introduces multiple prototypes clustering, which enables CLIP to learn curved decision boundaries while preserving its general knowledge.

In addition, we note that some works also employ probabilistic prompt learning for fine-tuning VLMs. For example,
ProDA~\cite{ProDA} proposes prompt distribution learning, which uses multiple prompts to construct a Gaussian distribution for each class and optimize it with surrogate loss.
DAPT~\cite{DAPT} optimizes both textual and visual prompts to adaptively learn suitable distributions within each modality.
APP~\cite{APP} introduces a data-dependent prior that allows prompts to better capture the complex distribution of image features.
Any-Shift~\cite{Any-shift} proposes a hierarchical probabilistic inference framework to generate pseudo-test prompts, explicitly modeling the relationship between training and test distributions.
Note that, existing works generally leverage probabilistic distributions for \emph{prompt} tuning. Differently, our approach leverages \textbf{graph-based distribution representation} in \emph{adapter}-based tuning methods by integrating the cues of both inter-class relationships and diverse/uncertain descriptions of each category together. 

\subsection{Graph-Based Representation Learning in VLMs}

Graph Neural Networks (GNNs) have shown superior capabilities in various computer vision tasks by modeling structural relationships~\cite{GCN,GAT}. 
Traditional GNNs rely on deterministic message propagation, where nodes are represented by fixed feature vectors. 
%
In the context of VLM adaptation, graph-based methods have shown promising results for exploring structural knowledge. 
For example, GraphAdapter~\cite{graphadapter} constructs a class-level dual knowledge graph using textual and visual features and applies the standard Graph Convolutional Networks (GCN)~\cite{GCN} to refine class embeddings. 
HeGraphAdapter~\cite{HeGraphAdapter} constructs a multi-modal heterogeneous graph that simultaneously models modality and class structure knowledge, and employs a Heterogeneous Graph Neural Network (HGNN) to extract task-specific structural knowledge.
Recent efforts~\cite{ZLaP,ECALP} propose to construct a graph consisting of text and images, followed by label propagation over the graph to improve zero-shot recognition. 
In contrast, for the first time, our work attempts to leverage a \textbf{random graph model}~\cite{beer2011vertex} to fine-tune VLM on downstream tasks which can leverage significantly more semantic knowledge in the adapter. 
%

\begin{figure*}
    \centering
    \includegraphics[width=1\textwidth]{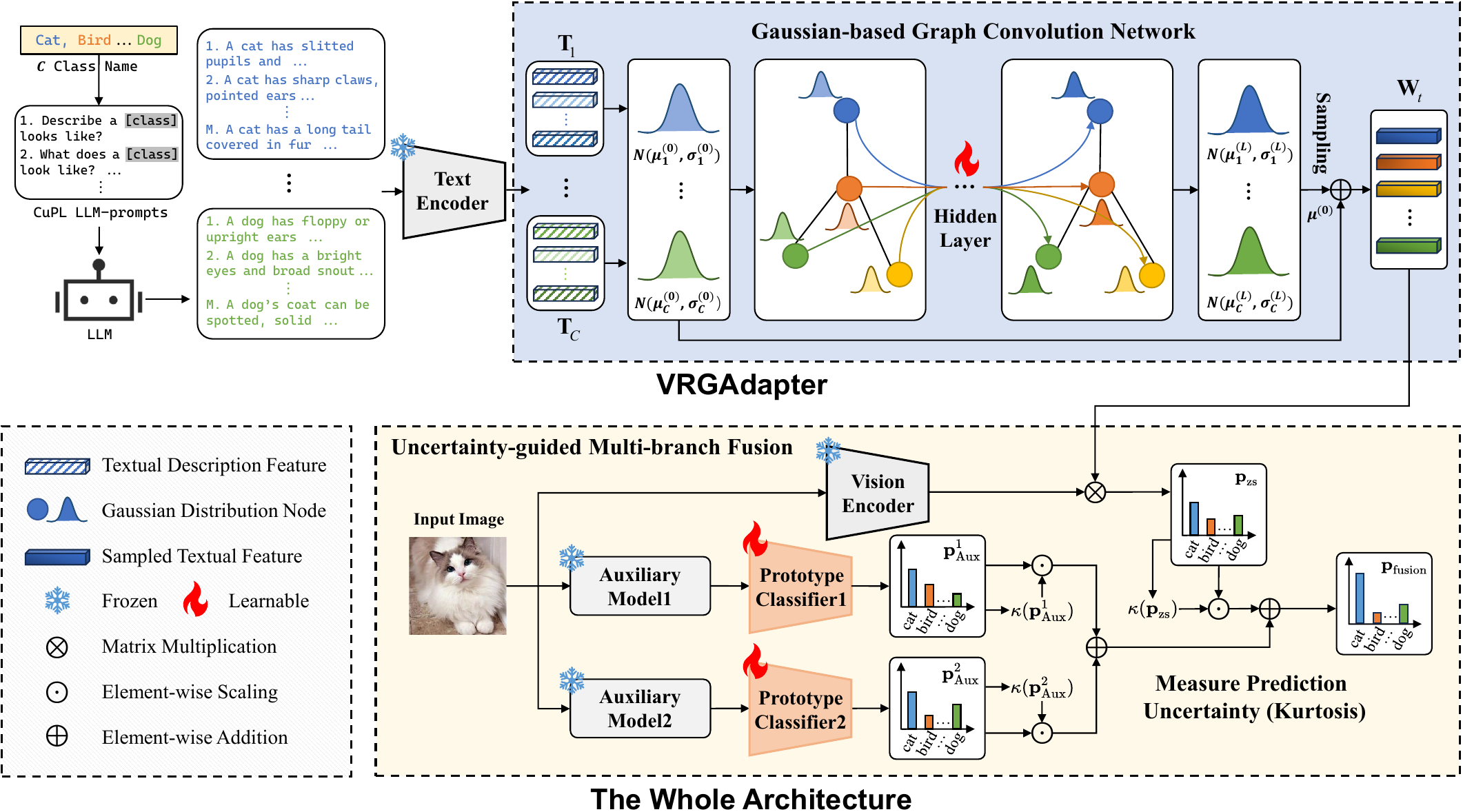}
    \caption{Overview of our proposed approach. The upper part illustrates the Vertex Random Graph Adapter (VRGAdapter) process: diverse descriptions are generated for each class, class-specific Gaussian distributions are computed to construct a Vertex Random Knowledge Graph (VRKG), probabilistic message propagation is performed on the VRKG to obtain context-aware class representations, and finally adaptive textual prototypes $\mathbf{W}_t$ are sampled from the refined distributions. 
    The lower part illustrates the Uncertainty-guided Multi-branch Fusion (UMF) process: multiple foundation models process the input image in parallel, and their predictions dynamically fused based on uncertainty measurements to produce the final classification result. 
    }
    \label{fig:overview}
\end{figure*}
\section{Vertex Random Graph Adapter}

In this section, 
we present our Vertex Random Graph Adapter (VRGAdapter) for fine-tuning VLMs. 
The aim of VRGAdapter is to fully capture both semantic diversity of each class and inter-class relationships of different classes to adapt the textual representation for the downstream tasks. 
Specifically, as illustrated in Fig.~\ref{fig:overview}, our proposed VRGAdapter contains three main components, i.e., Building Vertex Random Knowledge Graph (VRKG), Message Propagation on VRKG, and Reparameterized Sampling for Textual Adaptation. The details of these components are introduced in the following subsections respectively.  

\subsection{Vertex Random Knowledge Graph Building}

Given a dataset with $C$ classes, we can generate diverse and rich semantic descriptions for each class by following CuPL~\cite{cupl}, enabling better capturing of semantic variability within each class~\cite{cafo,cupl,AWT}. Specifically, these descriptions are first obtained via LLM~\cite{gpt3} as follows:
\begin{equation}
\label{eq:class_description}
[\mathbf{P}_i^1, \mathbf{P}_i^2, 
 \ldots,\mathbf{P}_i^M] = \text{LLM}(\text{``Describe class } i \text{''}),
\end{equation}
where $i \in [1,C]$ and $M$ denotes the number of textual descriptions generated per class. 
To encode textual information, the CLIP text encoder~\cite{clip} is then employed to yield semantic feature representations as:
\begin{equation}
\label{eq:text_encoding}
\mathbf{T}_i^m = f_T(\mathbf{P}_i^m),
\end{equation}
where $f_T$ denotes the CLIP text encoder. 
Our VRKG is defined as $\mathcal{G} = (\mathcal{V}, \mathcal{E})$, where 
 $\mathcal{V} = \{v_1, v_2, \ldots, v_C\}$ represents the set of class nodes and $\mathcal{E}$ represents the set of edges. 
 
\textbf{Nodes.} 
Unlike traditional knowledge graphs that represent each class using a deterministic feature vector~\cite{graphadapter,HeGraphAdapter}, we model each class node as a probabilistic distribution to fully capture the inherent semantic diversity within each category. Specifically, considering the diversity of class descriptions generated by LLM (Eq.~\ref{eq:class_description}), we model the  representation $\mathcal{H}_i$  for node $v_i$ in the VRKG as a Gaussian distribution:
\begin{equation}
\label{eq:Gaussian_distribution}
\mathcal{H}_i \sim \mathcal{N}\big(\boldsymbol{\mu}_i, \text{diag}(\boldsymbol{\sigma}_i)\big), \quad i \in [1,C],
\end{equation}
where $\boldsymbol{\mu}_i$ is the mean vector representing the semantic center of class $i$, and $\text{diag}(\boldsymbol{\sigma}_i)$ denotes the diagonal covariance matrix reflecting the semantic variability along each feature dimension.  
The distribution parameters are initially estimated based on the set of the above $M$ textual description features $\{\mathbf{T}_i^1, \mathbf{T}_i^2, \ldots, \mathbf{T}_i^M\}$ as follows:
\begin{align}
\label{eq:gaussian_distribution}
 &\boldsymbol{\mu}_i = \frac{1}{M} \sum\nolimits_{m=1}^M \mathbf{T}_i^m, \\ 
& \boldsymbol{\sigma}_i = \frac{1}{M} \sum\nolimits_{m=1}^M (\mathbf{T}_i^m- \boldsymbol{\mu}_i) \odot (\mathbf{T}_i^m - \boldsymbol{\mu}_i), 
\end{align}
where $\odot$ denotes element-wise multiplication operation.

\textbf{Edges.} The edges of VRKG encode the relationships between different classes. 
One straightforward way is to define the edge weight $\mathbf{A}_{ij}$ as the similarity (or correlation) between node representations $\mathcal{H}_i$ and $\mathcal{H}_j$. 
In this case, the obtained $\mathbf{A}_{ij}$ is probabilistic distribution because both $\mathcal{H}_i$ and $\mathcal{H}_j$ are Gaussian distributions. 
However, it is usually complex to define the message passing on the random graph with probabilistic edge weights. 
Therefore, to simplify, in this paper, we compute the edge weight $\mathbf{A}_{ij}$ of edge $e_{ij}$ based on the similarity between the mean representation of node $v_i$ and $v_j$, i.e., 
\begin{equation}
\label{eq:cosine_similarity}
\mathbf{A}_{ij} = \cos(\boldsymbol{\mu}_i, \boldsymbol{\mu}_j), \quad i,j \in [1,C],
\end{equation}
where $\boldsymbol{\mu}_i, \boldsymbol{\mu}_j$ denote the mean representation for nodes $v_i$ and $v_j$, respectively. 
Note that, since the edge weight $\mathbf{A}_{ij}$ is deterministic, we call the above graph $\mathcal{G}(\mathcal{H},\mathbf{A})$ as Vertex Random Knowledge Graph (VRKG).  

\subsection{Message Propagation on VRKG}

Once the VRKG $\mathcal{G}(\mathcal{H},\mathbf{A})$ is constructed, similar to 
regular graph neural network learning model~\cite{GCN,GAT}, 
we can learn context-aware representation for each class node by aggregating the information from other class nodes. 
One straightforward way is to apply Graph Convolutional Network (GCN)~\cite{GCN}. 
However, in our VRKG framework, each node is represented by a Gaussian distribution. Consequently, traditional graph convolution operations are not directly applicable. 
To overcome this, we employ the Gaussian-based graph convolution network~\cite{rgcn} to perform convolution operations on VRKG. 
It ensures that the graph node representation retains its Gaussian distribution after message propagation. 
Specifically, the message propagation on VRKG is conducted as: 
\begin{align}
\label{eq:UGC}
 \boldsymbol{\mu}_i^{(l+1)} &= \sigma\Big(\sum_{j \in \mathcal{N}(i)}\big(\mathbf{D}^{-1/2}\mathbf{A}\mathbf{D}^{-1/2}\big)_{ij}
\boldsymbol{\mu}_j^{(l)} \mathbf{\Theta}_\mathbf{\mu}^{(l)}\Big), \\
 \boldsymbol{\sigma}_i^{(l+1)} &= \sigma\Big(\sum_{j \in \mathcal{N}(i)}
\big(\mathbf{D}^{-1/2}\mathbf{A}\mathbf{D}^{-1/2}\big)_{ij}
\boldsymbol{\sigma}_j^{(l)} \mathbf{\Theta}_\sigma^{(l)}\Big),
\end{align}
where $l=0\ldots L-1$ denotes the layer index and the initial representations are set as $\boldsymbol{\mu}_i^{(0)} =\boldsymbol{\mu}_i$, $\boldsymbol{\sigma}_i^{(0)}=\boldsymbol{\sigma}_i$. $\mathbf{D} \in \mathbb{R}^{C \times C}$ denotes the diagonal matrix with $\mathbf{D}_{ii} = \sum_j \mathbf{A}_{ij}$. $\mathcal{N}(i)$ denotes the neighbors of class node $v_i$ including itself. $\mathbf{\Theta}_\mathbf{\mu}^{(l)}$ and $\mathbf{\Theta}_\sigma^{(l)}$ are learnable  matrices.   $\sigma(\cdot)$ is a nonlinear activation function.

\subsection{Reparameterized Sampling}
After $L$ layers of the above Gaussian-based graph convolution operation, we obtain refined Gaussian distribution for each class node by integrating the information of its neighbors.
To achieve textual feature adaptation, we sample from the final distributions as follows:
\begin{equation}
\mathbf{z}_i \sim \mathcal{N}\big(\boldsymbol{\mu}_i^{(L)}, \text{diag}(\boldsymbol{\sigma}_i^{(L)})\big).
\end{equation}
Since the sampling operation is non-differentiable, we adopt the reparameterization trick~\cite{kingma2013auto} to enable gradient-based optimization of the learnable parameters:
\begin{equation}
\label{eq:sampling}
\mathbf{z}_i = \boldsymbol{\mu}_i^{(L)} + \boldsymbol{\epsilon} \odot \sqrt{\boldsymbol{\sigma}_i^{(L)}}, \quad \boldsymbol{\epsilon} \sim \mathcal{N}(0, I),
\end{equation}
where $\boldsymbol{\epsilon}$ is a vector of random variables that follows a standard normal distribution and $\odot$ denotes element-wise multiplication operation.

To preserve the original semantic knowledge while incorporating adapter-enhanced information, we employ a weighted combination strategy:
\begin{equation}
\label{eq:residual_textual}
\mathbf{w}_i = \alpha\boldsymbol{\mu}_i + (1-\alpha)\mathbf{z}_i,
\end{equation}
where $\alpha \in [0,1]$ controls the trade-off between the initial class representation and the adapter-enhanced feature. 
In the following, we let $\mathbf{W}_t = \{\mathbf{w}_1, \mathbf{w}_2 \cdots \mathbf{w}_C\} \in \mathbb{R}^{C \times D}$ denote the set of final textual representations for $C$ classes after the VRGAdapter module. 
Notably, different from previous works,  $\mathbf{W}_t$ effectively leverages rich semantic diversity across LLM-generated descriptions to model uncertainty, providing enhanced representations for the downstream visual tasks.

\section{Uncertainty-guided Multi-branch Fusion}

\subsection{Multi-branch Architecture}

Recent works~\cite{cafo,amu} have demonstrated the benefits of integrating multiple pre-trained models like MoCo~\cite{mocov3} and DINO~\cite{dino} to enhance the performance for downstream tasks. In this section, we introduce a novel Uncertainty-guided Multi-branch Fusion (UMF) scheme that dynamically combines predictions from multiple foundation models based on their individual confidence levels.
To be specific, our framework operates in a $C$-way $N$-shot setting and consists of three complementary branches: one primary CLIP branch for general knowledge and two auxiliary branches for task-specific adaptation. 

\textbf{CLIP Encoder Branch.} To leverage the powerful zero-shot capabilities of pre-trained CLIP~\cite{clip}, we employ the frozen CLIP visual encoder to extract task-agnostic visual features. Given an input image $\mathbf{x}$, we obtain:
\begin{align}
&\mathbf{f}_{\text{clip}} = f_{V}(\mathbf{x}), \\
 &\mathbf{p}_{\text{zs}}= \mathbf{f}_{\text{clip}} \mathbf{W}_t^T,
\label{eq:clip_prediction}
\end{align}
where $f_V$ denotes the pre-trained CLIP visual encoder, $\mathbf{f}_{\text{clip}} \in \mathbb{R}^{1 \times D}$ represents the extracted visual features, $\mathbf{W}_t$ denotes the adaptive textual prototypes obtained from our VRGAdapter (Eq.~\ref{eq:residual_textual}) and $\mathbf{p}_{\text{zs}} \in \mathbb{R}^C$ denotes the zero-shot prediction scores.

\textbf{Auxiliary Model Branches.} To introduce task-specific visual knowledge while retaining CLIP's generalization capability, we incorporate two auxiliary foundation models $f_{\text{Aux}}^1$ and $f_{\text{Aux}}^2$ (e.g., MoCo~\cite{mocov3} and DINO~\cite{dino}). These auxiliary models provide complementary visual features that capture different aspects of visual samples.
To be specific, given the training set $\mathcal{T} = \{(\mathbf{x}_{ij}, y_i)\}_{i=1,j=1}^{C,N}$, where $C$ denotes the number of classes and $N$ represents the number of examples per class, we first construct prototype classifiers by averaging the features of training samples as follows:
\begin{align}
\label{eq:prototype_classifier}
&\mathbf{c}_i^k = \frac{1}{N}\sum_{j=1}^N f_{\text{Aux}}^k(\mathbf{x}_{ij}), \quad i\in [1,C], \\
&\mathbf{W}_{\text{proto}}^k = [\mathbf{c}_1^k, \mathbf{c}_2^k,\cdots,\mathbf{c}_C^k]^T , \quad k \in \{1,2\}, 
\end{align}
where $\mathbf{c}_i^k \in \mathbb{R}^D$ represents the prototype feature of class $i$ from the $k$-th auxiliary model, and $\mathbf{W}_{\text{proto}}^k \in \mathbb{R}^{C \times D}$ is the corresponding prototype classifier.

To enhance task-specific adaptation, for the image $\mathbf{x}$, these auxiliary models generate predictions by combining the prototype classifiers with learnable residual weights:
\begin{align}
\label{eq:auxiliary_prediction}
\mathbf{p}_{\text{Aux}}^k= {f}_{\text{Aux}}^k(\mathbf{x}) \big(\mathbf{W}_{\text{proto}}^k + \mathbf{W}_{\Delta}^k\big)^T,
\end{align}
where 
$\mathbf{W}_{\Delta}^k \in \mathbb{R}^{C \times D}$ denotes the learnable residual weights for task-specific adaptation, and $\mathbf{p}_{\text{Aux}}^k \in \mathbb{R}^C$ represents the prediction logits from the $k$-th auxiliary model.

\subsection{Dynamic Fusion}
After obtaining predictions from the CLIP encoder and auxiliary model branches, we apply an uncertainty-guided fusion mechanism to dynamically estimate each model's contribution based on its prediction confidence.
To quantify prediction uncertainty, we adopt a normalized kurtosis metric~\cite{kurtosis}:
\begin{align}
\label{eq:kurtosis}
\kappa(\mathbf{p}) = \left[\mathbb{E}\left(\frac{\mathbf{p} - \mu_{\mathbf{p}}}{\sigma_{\mathbf{p}}}\right)^4\right]^\lambda,
\end{align}
where $\kappa(\mathbf{p})$ indicates the prediction confidence. 
$\mu_{\mathbf{p}}$ and $\sigma_{\mathbf{p}}$ denote the mean and standard deviation of logits $\mathbf{p} \in \mathbb{R}^C$, and $\lambda > 0$ is a scaling parameter that controls the sensitivity of the confidence measure.
To effectively integrate the predictions of multiple models, we design a dynamic fusion strategy that balances logits according to their confidence scores:
\begin{align}
\mathbf{p}_{\text{bias}} &= \sum_{k=1}^2 \kappa(\mathbf{p}_{\text{Aux}}^k ) \cdot \mathbf{p}_{\text{Aux}}^k , \\
\mathbf{p}_{\text{fusion}} &= {\kappa(\mathbf{p}_{\text{zs}})} \cdot \mathbf{p}_{\text{zs}} +\beta \cdot \mathbf{p}_{\text{bias}},
\label{eq:fusion}
\end{align}
where $\beta$ is a balancing parameter that adjusts the relative influence of CLIP and auxiliary models. 
This adaptive fusion mechanism enables the model to exploit complementary knowledge from multiple foundation models while remaining robust to uncertainty.
The final prediction probabilities are obtained by applying softmax normalization to the fused logits as:
\begin{align}
\hat{\mathbf{p}}_{\text{fusion}}^i = \frac{\exp(\mathbf{p}_{\text{fusion}}^i)}{\sum_{j=1}^C \exp(\mathbf{p}_{\text{fusion}}^j)}, \quad i \in [1,C].
\end{align}
We train the entire framework by minimizing the cross-entropy loss as follows: 
\begin{align}
\label{eq:cross_entropy}
\mathcal{L}_{\text{CE}} = -\sum_{i=1}^C y_i \log(\hat{\mathbf{p}}_{\text{fusion}}^i),
\end{align}
where $y_i$ is the one-hot ground-truth label for the $i$-th training sample.

\section{Experiments}\label{sec:exam}
\subsection{Experimental Settings}

\subsubsection{Datasets} 
Following previous works~\cite{tip,cafo,amu}, we conduct extensive experiments on 11 diverse downstream tasks: ImageNet-1K~\cite{imagenet}, StanfordCars~\cite{car}, Caltech101~\cite{caltech101}, UCF101~\cite{ucf101}, Flowers102~\cite{flower102}, Food101~\cite{food101}, DTD~\cite{dtd}, EuroSAT~\cite{eurosat}, FGVCAircraft~\cite{fvga}, OxfordPets~\cite{pet}, and SUN397~\cite{sun397}. These datasets span a wide range of visual domains and classification tasks. Among them, OxfordPets (37 dog and cat breeds), Food101 (101 food classes), StanfordCars (196 car classes), Flowers102 (102 flower species), and FGVCAircraft (100 aircraft models) datasets represent fine-grained classification tasks that require the model to distinguish between visually similar subclasses. EuroSAT (10 land use classes) focuses on remote sensing image classification, while DTD (47  texture classes) evaluates texture recognition capabilities. The remaining datasets: Caltech101 (101 object classes), UCF101 (101 action classes), and SUN397 (397 scene classes) cover general object recognition, action recognition, and scene understanding, respectively. ImageNet-1K serves as our primary benchmark with its comprehensive collection of 1,000 diverse object classes.
To assess the model's robustness to out-of-distribution shifts, we also conduct generalization experiments on ImageNet-V2~\cite{imagenet-v2} and ImageNet-Sketch~\cite{imagenet-s} datasets. ImageNet-V2 represents a natural distribution shift from ImageNet-1K, while ImageNet-Sketch introduces style variations by replacing natural images with sketch-style images, presenting more challenging evaluation scenarios.

\subsubsection{Implementation details}
The training process utilizes the AdamW~\cite{adamW} optimizer initialized with a learning rate of 0.001, coupled with cosine learning rate decay. We train the model for 50 epochs using a fixed batch size of 256. 
For the VRGAdapter, we follow the approach of CuPL~\cite{cupl} to generate $M=50$ diverse textual descriptions per class.
We then apply a two-layer Gaussian-based graph convolution network ($L=2$), with the hidden feature dimension set to 16, and the residual weight $\alpha$ set to 0.7.
The activation function $\sigma(\cdot)$ is implemented as ELU~\cite{ELU} for mean vectors and ReLU~\cite{ReLU} for variance vectors.
The hyperparameters $\lambda$ in Eq.~\ref{eq:kurtosis} and $\beta$ in Eq.~\ref{eq:fusion} are tuned on the validation sets. All experiments are implemented in PyTorch and conducted on a single NVIDIA RTX 3090 GPU.
For the pre-trained models, we adopt three complementary pre-trained models as visual feature extractors.
Our primary model is CLIP which is implemented with the ResNet-50 architecture.
In addition, we employ two auxiliary models (i.e., MoCo~\cite{mocov3} and DINO~\cite{dino}), both also based on ResNet-50 for task-specific feature learning ($f_{\text{Aux}}^1$ and $f_{\text{Aux}}^2$).
We select MoCo and DINO because of their strong representation capabilities and widespread application in the vision community. 
Note that our uncertainty-aware fusion framework is model-agnostic and can readily incorporate other pre-trained models.

\begin{figure*}
    \centering
    \includegraphics[width=0.95\textwidth]{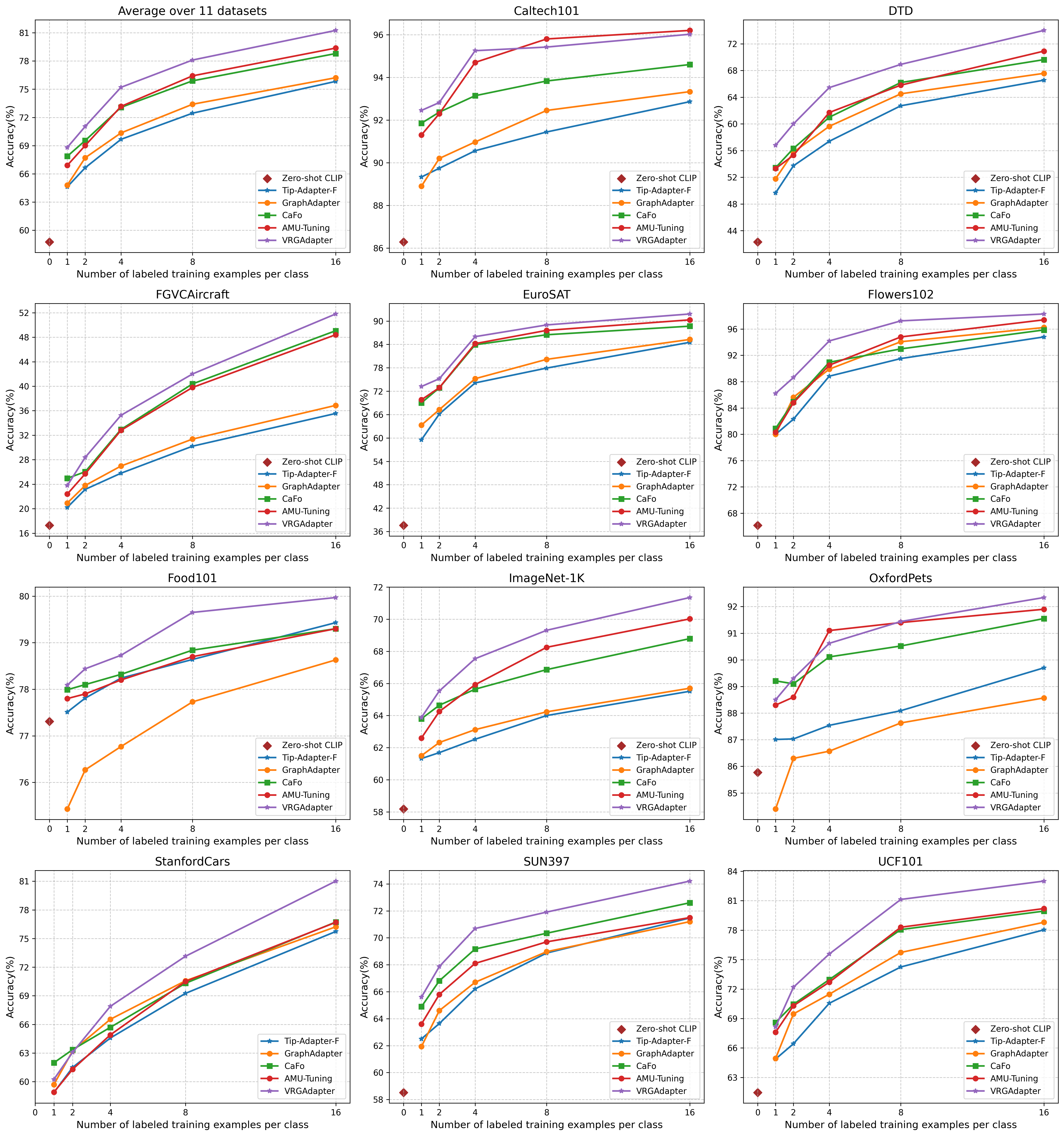}
    \caption{Performance comparison (\%) of different SOTA methods on few-shot classification, i.e., 1-/2-/4-/8-/16-shot, across 11 downstream tasks.
    }
    \label{fig:dataset11}
\end{figure*}

\begin{table}[!t]
\renewcommand{\arraystretch}{1.1}
\caption{Few-shot classification accuracy (\%) of different SOTA methods with ResNet50 backbone on ImageNet-1K.
}
\label{tab:imagenet}
\centering
\begin{tabular}{l|ccccc}
\hline
{\textbf{Method}}       & 1-shot     & 2-shot     & 4-shot     & 8-shot     & 16-shot \\
\hline
LP-CLIP~\cite{clip}     & 22.17      & 31.90      & 41.20      & 49.52      & 56.13 \\
CoOp~\cite{coop}        & 57.15      & 57.81      & 59.99      & 61.56      & 62.95 \\
ProDA~\cite{ProDA}      & 61.80      & 62.30      & 63.60      & 64.70      & 65.30 \\
CLIP-Adapter~\cite{clip-adapter} 
                        & 61.20      & 61.52      & 61.84      & 62.68      & 63.59 \\
Tip-Adapter-F~\cite{tip}& 61.32      & 61.69      & 62.52      & 64.00      & 65.51 \\
TaskRes~\cite{taskres}  & 61.43      & 62.17      & 62.93      & 64.03      & 65.75 \\
GraphAdapter~\cite{graphadapter} 
                        & 61.50      & 62.32      & 63.12      & 64.23      & 65.70 \\
ClusterAdapter~\cite{ClusterAdapter}& 62.60& 63.00& 63.80& 64.90&67.10\\
CaFo~\cite{cafo}        & 63.80      & 64.34      & 65.64      & 66.86      & 68.79 \\
DMN~\cite{dmn}          & 63.62      & 64.46      & 64.99      & 65.71      & 66.81 \\
AMU-Tuning~\cite{amu}   & 62.60      & 64.25      & 65.92      & 68.25      & 70.02 \\
VRGAdapter           & \textbf{63.88} & \textbf{65.53} & \textbf{67.54} & \textbf{69.31} & \textbf{71.35} \\
\hline
\end{tabular}
\end{table}

\begin{table*}[!t]
\renewcommand{\arraystretch}{1.1}
\caption{Comparison (\%) of different SOTA methods under various few-shot classification on 11 downstream tasks. The best performances are marked in bold.
}
\label{tab:dataset11}
\centering
\small
\begin{tabular}{c|c|ccccccccccc|c}
\hline
\textbf{Methods} & \textbf{Setting} & \rotatebox{90}{Caltech101} & \rotatebox{90}{DTD} & \rotatebox{90}{EuroSAT} & \rotatebox{90}{FGVCAircraft} & \rotatebox{90}{Flowers102} & \rotatebox{90}{Food101} & \rotatebox{90}{ImageNet-1K} & \rotatebox{90}{OxfordPets} & \rotatebox{90}{StanfordCars} & \rotatebox{90}{SUN397} & \rotatebox{90}{UCF101} & \textbf{Avg.} \\
\hline
 Zero-shot CLIP~\cite{clip}& 0-shot& 86.29& 42.32& 37.56& 17.28& 66.14& 77.31& 58.18& 85.77& 55.61& 58.52& 61.46&58.95\\
\hline
Tip-Adapter-F~\cite{tip} & \multirow{5}{*}{1-shot} & 89.33 & 49.65 & 59.53 & 20.22 & 79.98 & 77.51 & 61.32 & 87.01 & 58.86 & 62.50 & 64.87 & 64.62 \\
GraphAdapter~\cite{graphadapter} & & 88.90 & 51.77 & 63.30 & 20.93 & 80.00 & 75.43 & 61.50 & 84.40 & 59.70 & 61.93 & 64.93 & 64.80 \\
CaFo~\cite{cafo} & & 91.85 & 53.43 & 69.01 & \textbf{24.96} & 80.88 & 77.99 & 63.80 & \textbf{89.21} & \textbf{61.98} & 64.89 & \textbf{68.60} & 67.87 \\
AMU-Tuning~\cite{amu} & & 91.32 & 53.33 & 69.53 & 22.28 & 80.33 & 77.88 & 62.60 & 88.34 & 59.19 & 63.65 & 67.54 & 66.88 \\
VRGAdapter & & \textbf{92.45} & \textbf{56.80}& \textbf{73.20}& 23.79 & \textbf{86.20} & \textbf{78.09} & \textbf{63.88} & 88.50 & 60.24 & \textbf{65.60} & 68.15 & \textbf{68.81}\\
\hline
Tip-Adapter-F~\cite{tip} & \multirow{5}{*}{2-shot} & 89.74 & 53.72 & 66.15 & 23.16 & 82.30 & 77.81 & 61.69 & 87.03 & 61.50 & 63.64 & 66.43 & 66.65 \\
GraphAdapter~\cite{graphadapter} & & 90.20 & 55.75 & 67.27 & 23.80 & 85.63 & 76.27 & 62.32 & 86.30 & 63.23 & 64.60 & 69.47 & 67.71 \\
CaFo~\cite{cafo} & & 92.37 & 56.32 & 72.86 & 26.04 & 84.94 & 78.10 & 64.64 & 89.10 & \textbf{63.36} & 66.81 & 70.45 & 69.54 \\
AMU-Tuning~\cite{amu} & & 92.30 & 55.21 & 73.03 & 25.70 & 84.76 & 78.00 & 64.25 & 88.69 & 61.18 & 65.80 & 70.16 & 69.01 \\
VRGAdapter & & \textbf{92.82} & \textbf{59.99}& \textbf{75.20} & \textbf{28.38} & \textbf{88.63} & \textbf{78.44} & \textbf{65.53} & \textbf{89.30} & 63.08 & \textbf{67.87} & \textbf{72.19} & \textbf{71.04}\\
\hline
Tip-Adapter-F~\cite{tip} & \multirow{5}{*}{4-shot} & 90.56 & 57.39 & 74.12 & 25.80 & 88.83 & 78.24 & 62.52 & 87.54 & 64.57 & 66.21 & 70.55 & 69.67 \\
GraphAdapter~\cite{graphadapter} & & 90.97 & 59.63 & 75.20 & 26.97 & 89.90 & 76.77 & 63.12 & 86.57 & 66.53 & 66.70 & 71.47 & 70.35 \\
CaFo~\cite{cafo} & & 93.14 & 60.99 & 83.90 & 32.94 & 90.95 & 78.32 & 65.64 & 90.11 & 65.69 & 69.17 & 72.96 & 73.07 \\
AMU-Tuning~\cite{amu} & & 94.85 & 61.70 & 84.40 & 32.69 & 90.61 & 78.19 & 65.92 & \textbf{91.00} & 65.09 & 68.08 & 72.65 & 73.19 \\
VRGAdapter & & \textbf{95.25} & \textbf{65.43}& \textbf{85.99} & \textbf{35.25} & \textbf{94.19} & \textbf{78.73} & \textbf{67.54} & 90.62 & \textbf{67.89} & \textbf{70.69} & \textbf{75.57} & \textbf{75.20}\\
\hline
Tip-Adapter-F~\cite{tip} & \multirow{5}{*}{8-shot} & 91.44 & 62.71 & 77.93 & 30.21 & 91.51 & 78.64 & 64.00 & 88.09 & 69.25 & 68.87 & 74.25 & 72.45 \\
GraphAdapter~\cite{graphadapter} & & 92.45 & 64.50 & 80.17 & 31.37 & 94.07 & 77.73 & 64.23 & 87.63 & 70.57 & 68.97 & 75.73 & 73.40 \\
CaFo~\cite{cafo} & & 93.83 & 66.19 & 86.48 & 40.38 & 92.98 & 78.84 & 66.86 & 90.52 & 70.31 & 70.34 & 78.06 & 75.89 \\
AMU-Tuning~\cite{amu} & & \textbf{95.75} & 65.80 & 87.97 & 39.85 & 94.95 & 78.75 & 68.25 & 91.25 & 70.54 & 69.60 & 78.30 & 76.46 \\
VRGAdapter & & 95.42 & \textbf{68.91
}& \textbf{89.00} & \textbf{42.00} & \textbf{97.24} & \textbf{79.65} & \textbf{69.31} & \textbf{91.44} & \textbf{73.14} & \textbf{71.91} & \textbf{81.13} & \textbf{78.10}\\
\hline
Tip-Adapter-F~\cite{tip} & \multirow{5}{*}{16-shot} & 92.86 & 66.55 & 84.54 & 35.55 & 94.80 & 79.43 & 65.51 & 89.70 & 75.74 & 71.47 & 78.03 & 75.83 \\
GraphAdapter~\cite{graphadapter} & & 93.33 & 67.57 & 85.27 & 36.87 & 96.23 & 78.63 & 65.70 & 88.57 & 76.23 & 71.20 & 78.80 & 76.22 \\
CaFo~\cite{cafo} & & 94.60 & 69.62 & 88.68 & 49.05 & 95.86 & 79.30 & 68.79 & 91.55 & 76.73 & 72.60 & 79.94 & 78.79 \\
AMU-Tuning~\cite{amu} & & \textbf{96.28} & 70.82 & 90.22 & 48.39 & 97.58 & 79.28 & 70.02 & 92.01 & 76.72 & 71.40 & 80.30 & 79.37 \\
VRGAdapter & & 96.02 & \textbf{74.00}& \textbf{91.81} & \textbf{51.79} & \textbf{98.29} & \textbf{79.97} & \textbf{71.35} & \textbf{92.34} & \textbf{81.02} & \textbf{74.21} & \textbf{83.00} & \textbf{81.25}\\
\hline
\end{tabular}
\end{table*}

\subsection{Comparison with State-of-the-Art Methods}
We conduct comprehensive evaluations to demonstrate the effectiveness of our proposed method, comparing it with current state-of-the-art approaches on ImageNet-1K and 10 additional downstream tasks and out-of-distribution (OOD) benchmarks.

\subsubsection{Results on ImageNet-1K}
As shown in Table~\ref{tab:imagenet}, we compare our method with several strong baselines, including LP-CLIP~\cite{clip}, CoOp~\cite{coop}, ProDA~\cite{ProDA}, CLIP-Adapter~\cite{clip-adapter}, Tip-Adapter-F~\cite{tip}, TaskRes~\cite{taskres}, GraphAdapter~\cite{graphadapter}, ClusterAdapter~\cite{ClusterAdapter}, CaFo~\cite{cafo}, DMN~\cite{dmn} and AMU-Tuning~\cite{amu}. All methods use ResNet-50 as the backbone architecture for fair comparison.
Our method consistently outperforms all SOTA methods across different few-shot settings.
Specifically, compared with graph-based GraphAdapter~\cite{graphadapter} method, our model outperforms it by 2.38\%, 3.21\%, 4.42\%, 5.08\%, and 5.65\% under the 1-, 2-, 4-, 8-, and 16-shot settings, respectively. 
Compared with the AMU-Tuning~\cite{amu} method based on additional branch and logit bias, our model obtains the improvements of 1.28\%, 1.28\%, 1.62\%, 1.06\%, and 1.33\% under 1-, 2-, 4-, 8-, and 16-shot settings, respectively. 
For the 1-shot setting, our method slightly outperforms CaFo~\cite{cafo} and DMN~\cite{dmn}, which is especially significant considering that CaFo leverages additional synthetic training data generated by DALL-E~\cite{dalle} and DMN utilizes historical test data to assist classification. 
In contrast, our method achieves competitive performance using only the original few-shot training samples.
The larger improvements in higher-shot scenarios suggest that our method can effectively utilize and extract discriminative features from larger training sets.

\subsubsection{Results on Diverse Downstream Tasks} 
As illustrated in Fig.~\ref{fig:dataset11}, we evaluate our method across 11 diverse downstream tasks under various few-shot settings to thoroughly assess its robustness in different scenarios. For each dataset, we randomly sample the specified number of instances per class to form the training set, maintaining the original test splits for evaluation. The more detailed numerical results on downstream tasks can be found in Table~\ref{tab:dataset11}. Our proposed approach demonstrates superior performance across most tasks, showcasing its strong adaptability on diverse visual domains ranging from fine-grained classification to texture recognition and scene understanding. The average performance across all 11 downstream tasks demonstrates consistent improvements over state-of-the-art methods like GraphAdapter~\cite{graphadapter} and AMU-Tuning~\cite{amu}. These comprehensive results validate the effectiveness and strong generalization of our method.

\begin{table}[!htb]
\renewcommand{\arraystretch}{1.1}
\caption{Performance evaluation (\%) of different methods on out-of-distribution benchmarks.
}
\label{tab:Shift}
\centering
\begin{tabular}{l|c|cc}
\hline
\multirow{2}{*}{\textbf{Methods}} & \textbf{Source} & \multicolumn{2}{c}{\textbf{Target}} \\
\cline{2-4}
                               & ImageNet-1K        & -V2            & -Sketch        \\ 
\hline
\multicolumn{4}{c}{\textit{ResNet-50 Backbone}} \\
\hline
ZS-CLIP~\cite{clip}          & 58.18          & 51.34          & 33.32          \\
CoOp~\cite{coop}               & 62.95          & 55.11       & 32.74\\
CLIP-Adapter~\cite{clip-adapter}& 63.59          & 55.69          & 35.68          \\
Tip-Adapter-F~\cite{tip}       & 65.51          & 57.11          & 36.00          \\
TaskRes~\cite{taskres}         & 64.75          & 56.47          & 35.83          \\
GraphAdapter~\cite{graphadapter}& 64.94          & 56.58          & 35.89          \\
ClusterAdapter~\cite{ClusterAdapter}& 67.07          & 58.09          & 36.88          \\
CaFo~\cite{cafo}               & 68.79          & 57.99          & 39.43          \\
AMU-Tuning~\cite{amu}          & 70.02          & 58.64          & 40.04          \\
VRGAdapter                  & \textbf{71.35} & \textbf{62.41} & \textbf{41.23} \\
\hline
\multicolumn{4}{c}{\textit{ViT-B/16 Backbone}} \\
\hline
ZS-CLIP~\cite{clip}      & 66.73          & 60.83     & 46.15\\
CoCoOp~\cite{cocoop}           & 71.02          & 64.20          & 47.99          \\
MaPLe~\cite{maple}             & 70.72          & 64.07          & 49.15          \\
PromptSRC~\cite{PromptSRC}     & 71.27          & 64.35          & 49.55          \\
GraphAdapter~\cite{graphadapter}& 73.40          & 65.60          & 49.23          \\
AMU-Tuning~\cite{amu}          & 74.98          & 65.42          & 50.37          \\
LwEIB~\cite{LwEIB}        & 71.31          & 64.47          & 50.07          \\
MMRL~\cite{Mmrl}          & 72.03          & 64.47          & 49.17          \\
VRGAdapter                  & \textbf{76.78} & \textbf{68.60} & \textbf{51.78} \\
\hline
\end{tabular}
\end{table}

\subsubsection{Generalization to Out-of-Distribution Data}
To further evaluate the generalization capability of our method under distribution shifts, we conduct experiments on out-of-distribution (OOD) benchmarks by following previous works~\cite{coop,amu}. Specifically, we take models trained on ImageNet-1K under the 16-shot setting and directly evaluate them on two challenging variants: ImageNet-V2~\cite{imagenet-v2} and ImageNet-Sketch~\cite{imagenet-s}, representing natural distribution shift and style variation, respectively.
As shown in Table~\ref{tab:Shift}, our method achieves substantial improvements over the previous state-of-the-art methods. With the ResNet-50 backbone, our approach improves accuracy by 3.77\% on ImageNet-V2 and 1.19\% on ImageNet-Sketch compared to AMU-Tuning ~\cite{amu}. When using the ViT-B/16 backbone, our method provides additional gains of 3.18\% and 1.41\% on these two benchmarks, respectively. These results highlight the robustness and strong generalization ability of our VRGAdapter under natural distribution shifts.

\begin{table}[!t]
\renewcommand{\arraystretch}{1.1}
\caption{Ablation studies (\%)  of different components on ImageNet-1K. All experiments use ResNet-50 as the backbone.}
\label{tab:Component}
\centering
\setlength{\tabcolsep}{3.2pt}
\begin{tabular}{ccc|ccccc}
\hline
\multicolumn{3}{c|}{\textbf{Component}} & \multicolumn{5}{c}{\textbf{Accuracy (\%)}}\\
\hline
VRGAdapter & AUX & UMF & 1-shot & 2-shot & 4-shot & 8-shot & 16-shot \\
\hline
 &  &  & 59.42& 60.50& 61.69& 62.89& 64.46\\
\checkmark &  &  & 62.49 & 62.92& 63.57& 64.60& 66.03\\
 & \checkmark &  & 60.27& 62.53& 64.97& 67.47& 69.84\\
 & \checkmark & \checkmark & 60.77& 63.10& 65.48& 67.73& 70.05\\
\checkmark & \checkmark & \checkmark & \textbf{63.88} & \textbf{65.53} & \textbf{67.54} & \textbf{69.31} & \textbf{71.35} \\
\hline
\end{tabular}
\end{table}
\subsection{Ablation Studies}
We conduct extensive ablation experiments including the analysis of different components, hyperparameters, different CLIP visual encoders, computation efficiency, etc. More details of the analysis are as follows.

\subsubsection{Effectiveness of different components} 
To verify the effectiveness of each component in our framework, we conduct comprehensive ablation studies on ImageNet-1K~\cite{imagenet} under different few-shot settings. 
We use the prototype classifier of CLIP as the baseline for comparison. As shown in Table~\ref{tab:Component}, we investigate three key components: Vertex Random Graph Adapter (VRGAdapter), AUXiliary models (AUX), and Uncertainty-guided Multi-branch Fusion (UMF). 
First, employing the VRGAdapter module leads to consistent improvements across all settings, with accuracy increasing by 3.07\% in the 1-shot setting and 1.57\% in the 16-shot setting. These gains highlight the effectiveness of VRGAdapter in capturing intra-class semantic diversity and inter-class relationships.
Next, we incorporate auxiliary models and ensemble their predictions through averaging, which brings improvements across all few-shot settings. The performance gain becomes more obvious as the number of shots increases, from 0.85\% in 1-shot to 5.38\% in 16-shot, suggesting that auxiliary models can better leverage training samples to learn task-specific features that effectively complement CLIP's general visual knowledge.
Furthermore, adding the UMF scheme further improves performance consistently, with accuracy improvements of 0.50\% in the 1-shot setting and 0.57\% in the 2-shot setting. 
This validates the effectiveness of our uncertainty-based multi-branch fusion strategy in dynamically combining predictions from different models based on their confidence levels. 
Finally, when all components are integrated, our full model achieves substantial improvements over the baseline across all settings, with an increase of 4.46\% in the 1-shot setting and 6.89\% in the 16-shot setting.
These results clearly validate the complementary roles of our proposed components: VRGAdapter captures intra-class semantic diversity and inter-class relationships in the textual features; AUX introduces task-adaptive visual features via auxiliary pre-trained models; and UMF dynamically integrates model predictions based on confidence.

\begin{table*}[!t]
\renewcommand{\arraystretch}{1.1}
\caption{Comparison (\%)  between GraphAdapter and our VRGAdapter on DTD and ImageNet-1K under various few-shot settings based on ResNet-50 architecture. }
\label{tab:vsGraphAdapter}
\centering
\begin{tabular}{l|ccccc|ccccc}
\hline
\multirow{2}{*}{\textbf{Method}} & \multicolumn{5}{c|}{\textbf{DTD}} & \multicolumn{5}{c}{\textbf{ImageNet-1K}}\\
\cline{2-11}
 & 1-shot & 2-shot & 4-shot & 8-shot & 16-shot & 1-shot & 2-shot & 4-shot & 8-shot & 16-shot \\
\hline
CLIP + MoCo + DINO & 52.54& 53.61& 61.11& 65.07& 69.68& 60.77& 63.10& 65.48& 67.73& 70.05\\
 + GraphAdapter & 54.08& 56.15& 63.89& 67.73& 72.87& 62.12& 64.12& 66.70& 68.77& 70.83\\
 + VRGAdapter & \textbf{56.80}& \textbf{59.99}& \textbf{65.43}& \textbf{68.91}& \textbf{74.00}& \textbf{63.88} & \textbf{65.53} & \textbf{67.54} & \textbf{69.31} & \textbf{71.35} \\
\hline
\end{tabular}
\end{table*}

\subsubsection{Comparison with GraphAdapter}
To further validate the effectiveness of our proposed VRGAdapter, we conduct additional experiments comparing it against GraphAdapter~\cite{graphadapter} with the same settings. 
Note that our baseline integrates MoCo~\cite{mocov3} and DINO~\cite{dino}, both based on ResNet-50 architecture, as two auxiliary models to assist CLIP classification via the proposed UMF strategy.
As shown in Table~\ref{tab:vsGraphAdapter}, the results on both DTD~\cite{dtd} and ImageNet-1K~\cite{imagenet} datasets demonstrate several key findings: 
(1) GraphAdapter enhances performance by modeling class relationships through deterministic message propagation, achieving improvements of 1.54\%–3.19\% on DTD and 0.78\%–1.35\% on ImageNet-1K across different few-shot settings.
(2) Our VRGAdapter achieves substantial improvements over the baseline, with gains of 3.84\%–6.38\% on DTD and 1.30\%–3.11\% on ImageNet-1K across all settings. Moreover, it consistently outperforms GraphAdapter across different few-shot settings, with relative gains of 1.13\%–3.84\% on DTD and 0.52\%–1.76\% on ImageNet-1K.
The superior performance of VRGAdapter can be attributed to its explicit modeling of diversity in class representations and inter-class relationships. Unlike GraphAdapter, which treats class nodes as deterministic vectors, VRGAdapter captures the inherent semantic diversity within each class and models the inter-class relationships via probabilistic message propagation.


\begin{figure}
    \centering
    \includegraphics[width=0.45\textwidth]{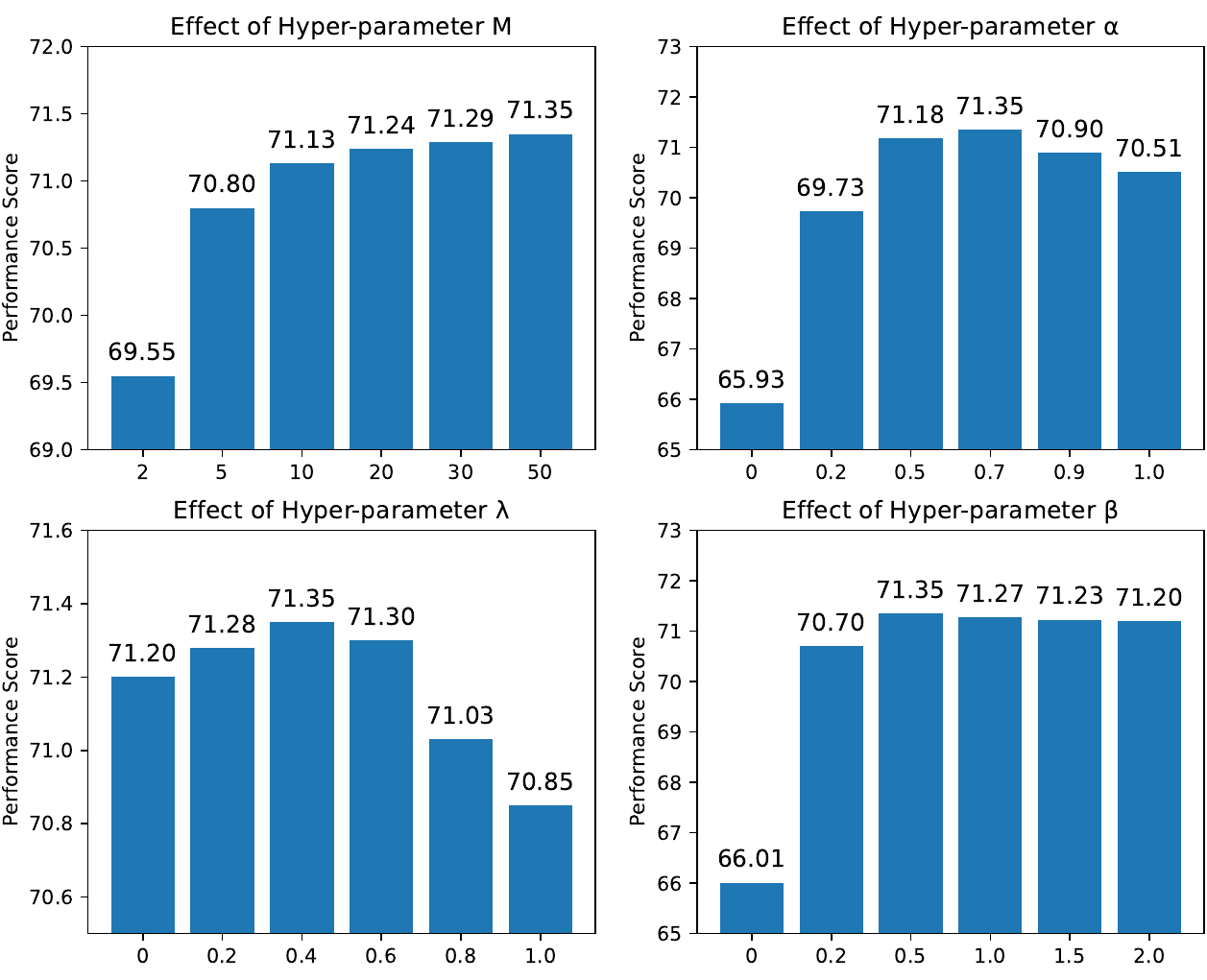}
    \caption{Ablation studies (\%) for hyperparameters on ImageNet-1K. 
    $M$: number of descriptions per class; $\alpha$: trade-off parameter in VRGAdapter; $\lambda$: uncertainty scale in Kurtosis; $\beta$: fusion weight in multi-model fusion. The experiments are conducted under the 16-shot setting with the ResNet-50 backbone.}
    \label{fig:parameter}
\end{figure}

\subsubsection{Impact of Hyperparameters}
We investigate the sensitivity of four main hyperparameters in our framework on ImageNet-1K~\cite{imagenet} under 16-shot setting, including the number of descriptions $M$ (Eq.~\ref{eq:class_description}), trade-off parameter $\alpha$ (Eq.~\ref{eq:residual_textual}), uncertainty scale $\lambda$ (Eq.~\ref{eq:kurtosis}), and fusion weight $\beta$ (Eq.~\ref{eq:fusion}). As shown in Fig.~\ref{fig:parameter}, we conduct extensive experiments to analyze these parameters.
For the number of descriptions $M$, we observe that the performance consistently improves as $M$ increases from 2 to 50, with the best accuracy of 71.35\% achieved at $M=50$. This indicates that generating more diverse descriptions helps capture richer semantic information of each class. However, the improvement gradually saturates when $M$ exceeds 20, suggesting that a moderate number of descriptions is sufficient for stable performance.
The trade-off parameter $\alpha$ controls the balance between the initial and adapter-enhanced representations in VRGAdapter. When $\alpha = 0$, the model only uses adapter-enhanced features, resulting in a poor performance of 65.93\%. 
Our model achieves the optimal balance when we set $\alpha$ to 0.7, while further increasing $\alpha$ leads to performance degradation. This demonstrates the importance of properly combining initial semantic features and adapter-enhanced information.
The uncertainty scale $\lambda$ affects the sensitivity of confidence measurement in Kurtosis calculation. We find that a moderate value of $\lambda=0.4$ works best. Either too small or too large values of $\lambda$ lead to suboptimal results, as they may under- or over-emphasize the uncertainty in model predictions.
The fusion weight $\beta$ balances the contributions between CLIP and auxiliary models. The performance is relatively poor without fusion, and setting $\beta=0.5$ achieves the best results, while larger values show slight degradation. This suggests that properly weighting different model predictions is crucial for an effective ensemble.

\begin{figure*}[t]
    \centering
    \begin{tabular}{ccc}
        \includegraphics[width=0.31\linewidth]{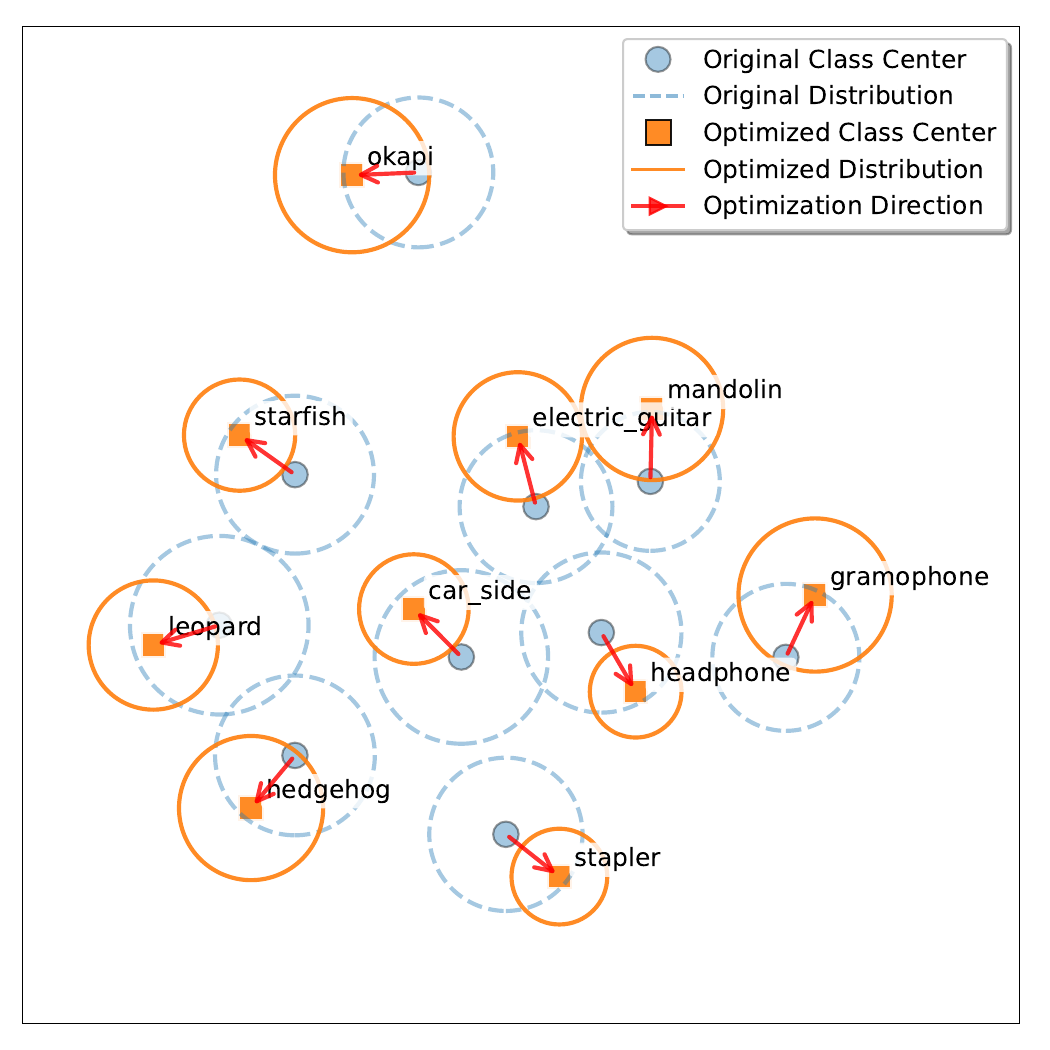} &
        \includegraphics[width=0.31\linewidth]{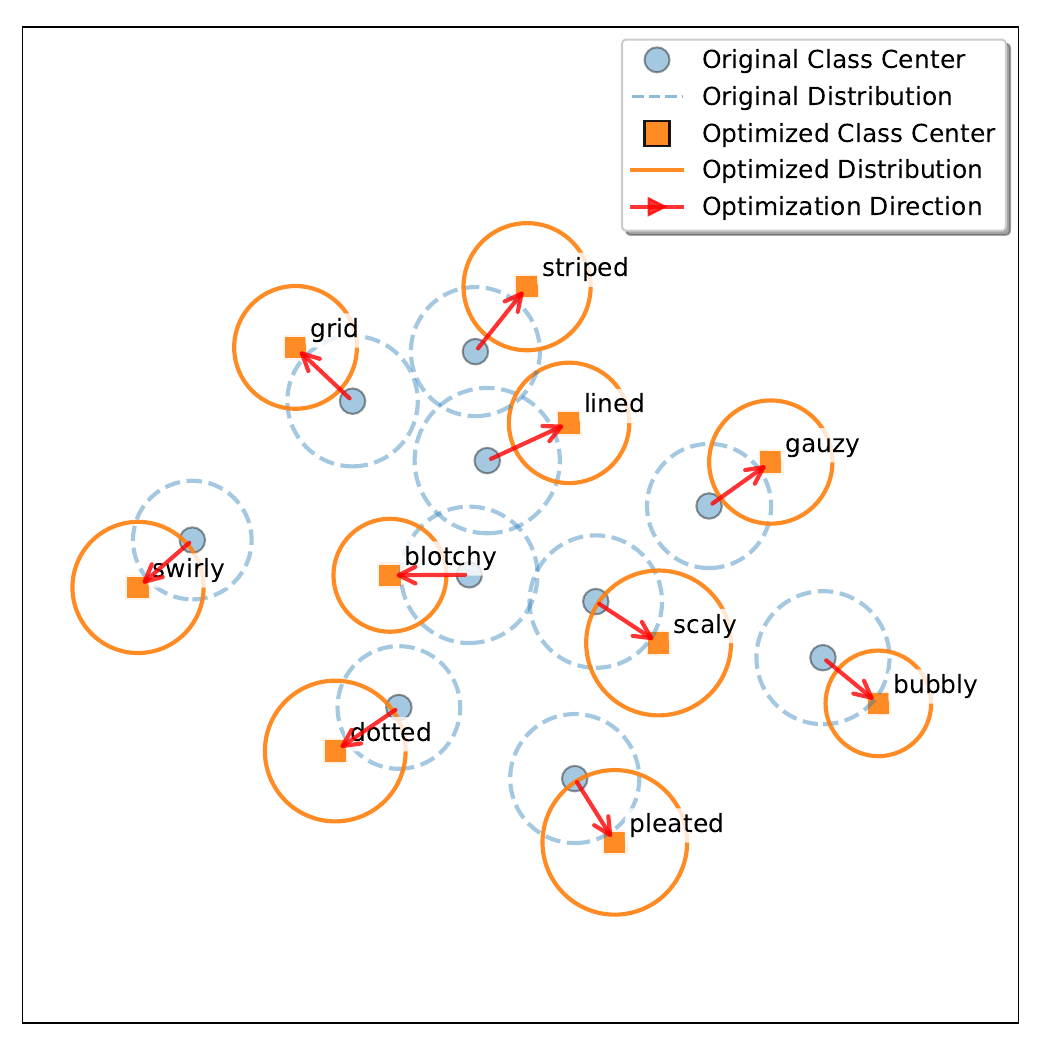} &
        \includegraphics[width=0.31\linewidth]{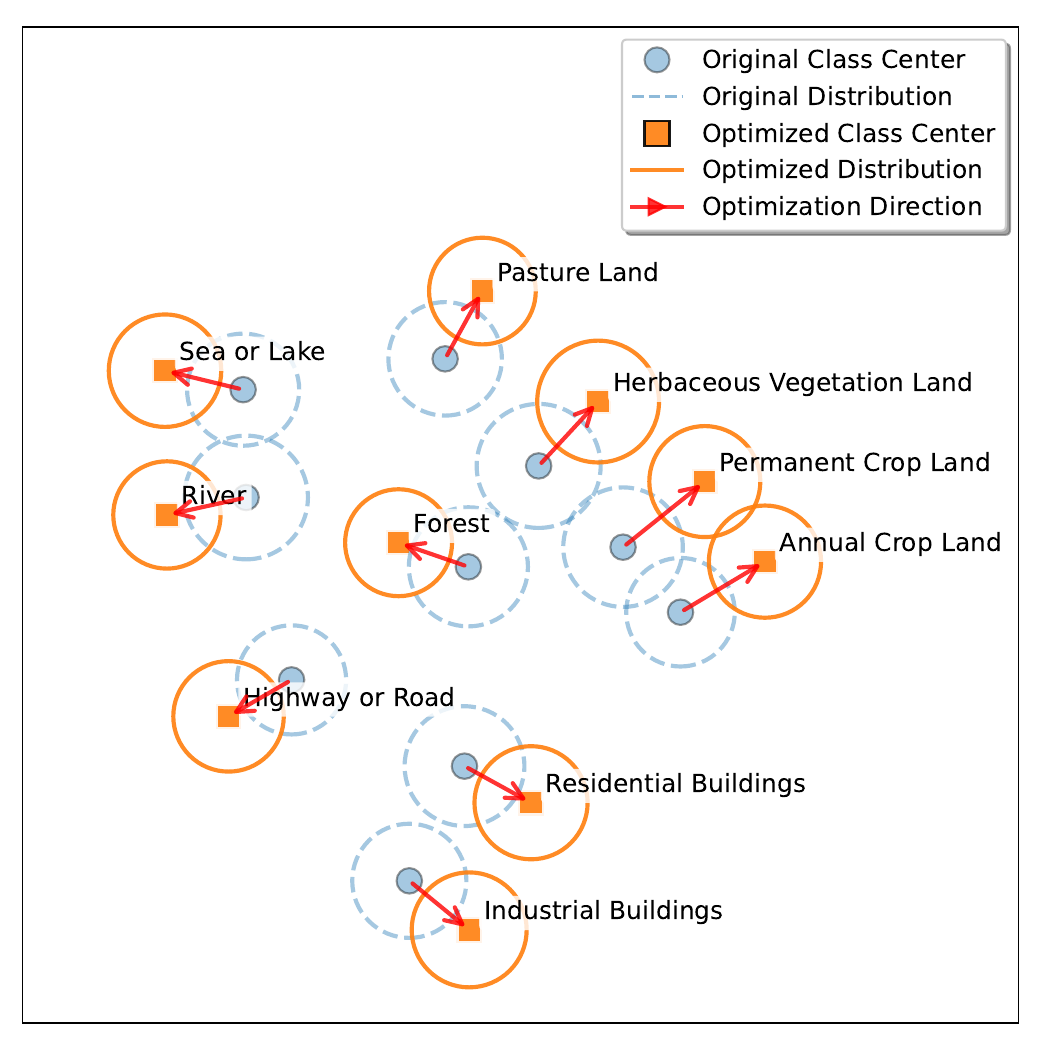} \\
        (a) Caltech101  & (b) DTD & (c) EuroSAT \\
    \end{tabular}
    \caption{2D t-SNE visualization of class distributions before and after applying VRGAdapter. Original class centers (circles) and optimized class centers (squares) are shown for 10 randomly selected classes. \textcolor{red}{Red} arrows indicate optimization directions, while dashed and solid circles represent distribution variance ranges before and after optimization, respectively.}
    \label{fig:visualization}
\end{figure*}

\subsubsection{CLIP Visual Encoders}
Furthermore, we evaluate our method on ImageNet-1K~\cite{imagenet}  dataset with various CLIP visual encoders, including CNN-based (ResNet-50~\cite{resnet}, ResNet-101~\cite{resnet}) and transformer-based (ViT-B/32~\cite{vit}, ViT-B/16~\cite{vit}) architectures. As shown in Table~\ref{tab:encoder}, our method consistently demonstrates superior performance compared to existing approaches across all backbone variants, validating the generalization ability of our proposed method.

\begin{table}[!t]
\renewcommand{\arraystretch}{1.1}
\caption{Ablation Study (\%) of different CLIP visual encoders on 16-shot ImageNet-1K.
}
\label{tab:encoder}
\centering
\begin{tabular}{l|cccc}
\hline
\multirow{2}{*}{\textbf{Models}} & \multicolumn{4}{c}{\textbf{Backbone}} \\
\cline{2-5}
                       & RN50  & RN101 & ViT-B/32 & ViT-B/16 \\
\hline
ZS-CLIP~\cite{clip}           & 60.33 & 65.53 & 63.80  & 68.73 \\
CoOp~\cite{coop}              & 62.95 & 66.60 & 66.85  & 71.92 \\
CLIP-Adapter~\cite{clip-adapter} & 63.59 & 65.39 & 66.19  & 71.13 \\
TaskRes~\cite{taskres}        & 64.75 & 67.70 & 68.20  & 73.07 \\
Tip-Adapter-F~\cite{tip}      & 65.51 & 68.56 & 68.65  & 73.69 \\
GraphAdapter~\cite{graphadapter} & 65.70 & 68.23 & 68.80 & 73.68 \\
ClusterAdapter~\cite{ClusterAdapter}& 67.07& 69.49& 69.74&74.52\\
CaFo~\cite{cafo}              & 68.79 & 70.82 & 70.82  & 74.48 \\
AMU-Tuning~\cite{amu}         & 70.02 & 71.58 & 71.65  & 74.98 \\
VRGAdapter                 & \textbf{71.35} & \textbf{73.28} & \textbf{73.34} & \textbf{76.78} \\
\hline
\end{tabular}
\end{table}

\subsubsection{Computation Efficiency}
We also compare the computational efficiency and model performance of different methods in Table~\ref{tab:efficiency}. All experiments are conducted on a single NVIDIA RTX 3090 GPU using the 16-shot ImageNet-1K dataset. As shown in the table, CoOp~\cite{coop} requires minimal parameters but requires substantial training time and computational resources for 200 epochs, as it performs gradient back-propagation through the entire textual encoder.
In contrast, methods like Tip-Adapter-F~\cite{tip} and CaFo~\cite{cafo} significantly reduce training time but at the cost of increased parameter count, particularly due to the fine-tuning of their cache models.
Although our method requires more learnable parameters than GraphAdapter~\cite{graphadapter} and AMU-Tuning~\cite{amu}, it achieves 5.65\% higher accuracy than GraphAdapter and 1.33\% higher accuracy than AMU-Tuning, respectively. This demonstrates an effective balance between computational efficiency and performance.

\begin{table}[!t]
\renewcommand{\arraystretch}{1.1}
\caption{Comparison of Accuracy (\%) and Efficiency on ImageNet-1K dataset under the 16-shot setting.}
\label{tab:efficiency}
\centering
\begin{tabular}{l|cccc}
\hline
\textbf{Methods} & \textbf{Training} & \textbf{Epochs} & \textbf{Param.} & \textbf{Acc.} \\
\hline
ZS-CLIP~\cite{clip} & - & - & - & 60.33 \\
CoOp~\cite{coop} & 14 h & 200 & 0.01 M & 62.95 \\
CLIP-Adapter~\cite{clip-adapter} & 50 min & 200 & 0.52 M & 63.59 \\
Tip-Adapter-F~\cite{tip} & 5 min & 20 & 16.3 M & 65.51 \\
GraphAdapter~\cite{graphadapter} & 40 min & 100 & 4.14 M & 65.70 \\
CaFo~\cite{cafo} & 20 min & 20 & 49.1 M & 68.79 \\
AMU-Tuning~\cite{amu} & 30 min & 50 & 2.05 M & 70.02 \\
VRGAdapter & {40 min} & {50} & {6.31 M} & {71.35} \\
\hline
\end{tabular}
\end{table}

\subsection{Visualization.} 
To visualize the effectiveness of our proposed VRGAdapter in modeling semantic diversity and refining inter-class relationships, we perform 2D t-SNE~\cite{tSNE} visualization of class distributions before and after applying our method, as shown in Fig.~\ref{fig:visualization}. 
We randomly select 10 classes from three different datasets, including Caltech101~\cite{caltech101}, DTD~\cite{dtd} and EuroSAT~\cite{eurosat}.
The visualization reveals three key insights:
(1) The optimized class centers (squares) exhibit enhanced inter-class separability compared to the original centers (circles). For example, in the DTD dataset, texture classes such as `blotchy', `lined’ and `scaly' demonstrate improved discrimination in the feature space.
(2) The comparison between original distributions (dashed circles) and optimized distributions (solid circles) reveals that some classes exhibit adaptive diversity adjustment. For example, in the Caltech101 dataset, well-defined classes such as `starfish' exhibit reduced diversity, while categories that are more difficult to define, such as `gramophone' appropriately maintain higher diversity to reflect their inherent visual variability.
These observations confirm that VRGAdapter effectively models both intra-class semantic diversity and inter-class relationships, yielding more reliable representations for the downstream applications.

\section{Conclusion}
In this paper, we propose VRGAdapter, a novel adapter framework that leverages random graph learning to jointly model the intra-class semantic diversity and inter-class relationships. Specifically, our approach introduces a Vertex Random Knowledge Graph (VRKG) where each node is represented as a Gaussian distribution estimated from diverse LLM-generated descriptions and edges encode semantic correlations between classes. Through probabilistic message propagation on this graph, our method effectively generates context-aware distribution representation for each class.
Additionally, we introduce an uncertainty-guided multi-branch fusion scheme that dynamically integrates predictions from multiple pre-trained visual encoders.
Extensive experiments on 11 benchmark datasets demonstrate substantial improvements over state-of-the-art methods on few-shot classification and out-of-distribution generalization.


\bibliographystyle{IEEEtran}
\bibliography{reference}

\end{document}